\documentclass[conference]{IEEEtran}
\usepackage{graphicx}
\usepackage{cite}
\usepackage{amsmath}
\usepackage{url}
\usepackage{caption}
\usepackage{subcaption}
\usepackage{hyperref}
\usepackage{orcidlink}

\begin{document}

\title{Low-Cost Neuromorphic Fall Detection Using Synthetic Event Data and Hybrid SNNs}

\author{\IEEEauthorblockN{Guillermo Rojas, Gonzalo Soto, Daniel Yunge\orcidlink{0000-0001-7149-2768}}
\IEEEauthorblockA{School of Electrical Engineering\\
Pontificia Universidad Católica de Valparaíso, Chile\\
guillermo.rojas.n@mail.pucv.cl, gonzalo.soto.c01@mail.pucv.cl, daniel.yunge@pucv.cl}
}

\maketitle

\begin{abstract}
This work presents the development of hybrid models that integrate spiking neural networks (SNNs) with components of convolutional neural networks (CNNs) to learn from simulated event-based camera data (Dynamic Vision Sensor, DVS) generated from conventional smartphone videos. Aimed primarily at human fall detection, the approach leverages the energy efficiency and spatio-temporal processing capabilities of SNNs by converting video frames into event-based data. The proposed models are evaluated through simulations on multiple datasets, comparing their performance to that of traditional machine learning models. Results demonstrate significant gains in efficiency without sacrificing accuracy, underscoring the potential of combining SNNs and DVS technology for complex tasks in real-world environments.
\end{abstract}


\section{Introduction}
In recent years, Spiking Neural Networks (SNNs) have gained attention as an energy-efficient alternative to traditional Artificial Neural Networks (ANNs), offering event-driven computation that closely mimics biological neurons. This efficiency is further amplified when combined with neuromorphic hardware, which supports parallel, low-power processing ideal for real-time applications. A critical enabler in such systems is the Dynamic Vision Sensor (DVS), an event-based camera that asynchronously captures changes in the visual scene with high temporal resolution and dynamic range \cite{gallego2020event}. The sparse, motion-focused data produced by DVS sensors makes them especially suitable for applications like robotics, surveillance, and wearable devices where low latency and power consumption are crucial.

Ambient-Assisted Living (AAL) is a particularly promising domain for the adoption of DVS cameras and neuromorphic computing. AAL systems aim to support independent living for the elderly and individuals with mobility limitations, often through activity monitoring and fall detection \cite{cicirelli2021ambient}. Fall detection methods are broadly categorized into wearable and ambient systems \cite{wang2020elderly,singh2020sensor}. Wearables, such as accelerometer- or gyroscope-based devices, are portable and low-cost but rely heavily on user compliance. Ambient systems, including those using RGB or depth cameras and radar, offer unobtrusive monitoring but often raise privacy concerns and require complex installation. In contrast, DVS cameras offer a compelling middle ground, preserving privacy while providing detailed motion information.

Recent research has demonstrated the potential of DVS sensors for fall detection and action recognition. \cite{wang2023fall} evaluated the accuracy and efficiency of Convolutional Neural Networks (CNNs), Long-Short Term Memory (LSTMs), and Recurrent SNNs using simulated DVS data by means of v2e \cite{hu2021v2e} over RGB videos from a body worn camera. They found that RSNNs, though slightly less accurate than CNNs, drastically reduced computational load. \cite{prasad2023hybrid} introduced DVSFall, a dataset collected with real DVS sensors, and demonstrated a 97.84\% accuracy using a hybrid 3D-CNN and SNN model. Similarly, \cite{lee2017embedded} proposed a real-time temporal network (DVS-TN) that achieved a 95.5\% F1-score on embedded hardware. Further, \cite{krishnan2022benchmarking} showed that conventional deep learning models like MViT could be adapted for DVS data, achieving up to 95.8\% accuracy. Finally, \cite{belbachir2012care} developed the CARE system, a stereo-DVS-based embedded solution that ensures privacy while delivering over 90\% detection accuracy in real-world scenarios. These studies collectively affirm the feasibility and advantages of DVS-based systems for reliable, efficient, and privacy-conscious fall detection.

Our work presents a low-cost framework based on \cite{wang2023fall} in which a hybrid combination of SNN and CNNs are trained offline using simulated event-based vision from smartphone videos to perform fall detection in an accurate, real-time, and low-power fashion. Whereas \cite{wang2023fall} used wearable cameras, down-sampled event data to a spatial resolution of 32x32, and trained 13 general activities of daily living (four fall types among them), we used static smartphone cameras, considered a spatial resolution of 128x128, and focused on three simple, yet similar activities of daily living to create our dataset, namely walking, sitting and falling. Also, an event-based dataset named NFDD was created for the model training of these activities. The proposed pipeline presented in this paper consists of four main sections: (1) Hybrid SNN-CNN model design and training, (2) neuromorphic dataset creation (NFDD), (3) model adaptation and fine-tuning, and (4) evaluation and comparative analysis.

The paper is organized as follows: Section II describes the methodology, Section III presents the experimental setup, datasets used, training protocol, and baseline comparisons. Section IV shows the evaluation metrics and experimental results. Finally, Section V discusses the conclusions and outlines directions for future work.

\section{Methodology}
As mentioned, the methodological framework of this research is grounded in the exploration of hybrid spiking neural networks (SNNs) combined with convolutional neural networks (CNNs), trained on neuromorphic data synthetically generated from conventional RGB video through the use of the v2e (video to events) simulation library. snnTorch was used as simulation framework, whereas the Tonic and Torch packages were used for the dataset processing. The project, called NFDD\_SNN, is available as a Kaggle repository \cite{nfdd2024}. The core objective was to construct a low-cost system capable of accurately detecting human falls using neuromorphic elements, while overcoming the accessibility limitations of real event-based cameras. Accordingly, we first created a preliminary architecture, and tested it with a standard dataset, to then create an application-specific dataset for further improvement of the architecture.

\subsection{DVSGesture}
\begin{figure}[t]
\centering
\includegraphics[width=\linewidth]{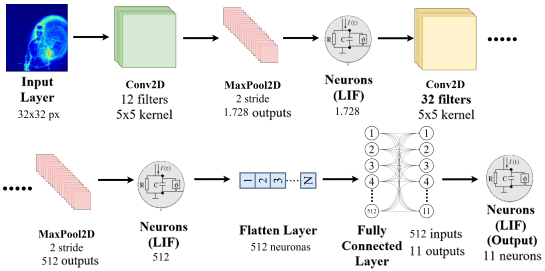}
\caption{Hybrid neural network designed for the DVSGesture dataset. 32x32px input, 1728 + 512 + 11 LIF neurons.}
\label{fig:1}
\end{figure}

\begin{figure}
\centering
\begin{subfigure}{.25\textwidth}
  \centering
\includegraphics[width=.95\linewidth]{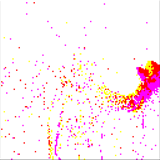}
  \caption{Original 128x128px}
  \label{fig:sub1}
\end{subfigure}%
\begin{subfigure}{.25\textwidth}
  \centering
\includegraphics[width=.95\linewidth]{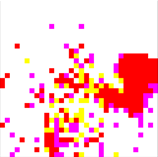}
  \caption{Downsampled 32x32px}
  \label{fig:sub2}
\end{subfigure}
\caption{Hundred events integrated into a 128x128px frame, and its downsampled version to 32x32px of a person moving the hand.}
\label{fig:2}
\end{figure}

Initially, a hybrid model architecture incorporating both CNN and SNN components was designed and evaluated using the public DVS128Gesture dataset \cite{amir2017low}. This dataset, acquired using real dynamic vision sensors (DVS), provided a strong baseline for evaluating the model’s ability to recognize gestures in an event-based setting. The original architecture is shown in Fig.\ref{fig:1}, and comprises four main stages: a) events from the DVS128Gesture dataset are integrated by 100 to create a 128x128 frame, which is later downsampled to 32x32 for efficiency reasons (Fig.\ref{fig:2}), b) a 2D convolutional filter set (12 filters, 5x5 kernel, 2 stride) connected to 1728 LIF neurons, c) another 2d convolutional filter set (32 filters, 5x5 kernel, 2 stride) connected to 512 LIF neurons, and d) a 512x11 fully connected layer to 11 LIF neurons, representing 11 classes, from which only three were used for the mentioned activities. Training was perform using the surrogate gradient descent, included in snnTorch, method and an Adam optimizer included in Torch.

\begin{figure}[t]
\centering
\includegraphics[width=\linewidth]{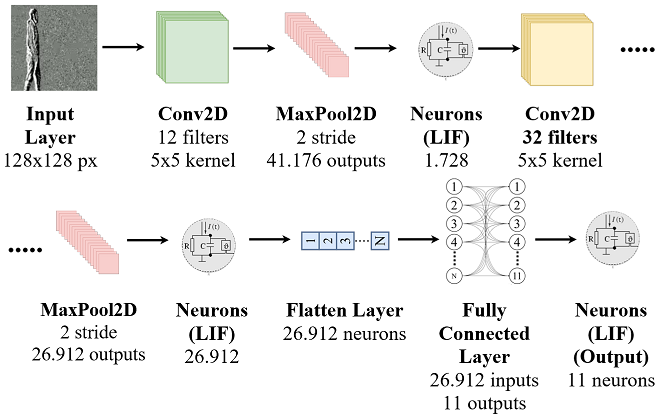}
\caption{Hybrid neural network designed for NFDD dataset. 128x128px input, 1728 + 26912 + 11 LIF neurons.}
\label{fig:3}
\end{figure}

Once a satisfactory accuracy of 91.4\% was obtained after 100 epochs (64 iterations each) on this benchmark dataset, a second, more complex architecture was proposed, focused specifically on fall detection, as shown in Fig.\ref{fig:3}.

\subsection{Neuromorphic Falling Detection Dataset (NFDD)}
This dataset, named the Neuromorphic Falling Detection Dataset (NFDD), was developed as a low-cost approach to train an SNN-based model to realistic environments without requiring physical DVS hardware. Unlike existing works which rely on custom camera systems or event cameras, this project leverages standard smartphone RGB cameras to capture raw video, which is then transformed into event data through the v2e simulation tool \cite{hu2021v2e}.
The v2e library plays a central role in the methodology, as it enables the generation of highly realistic synthetic DVS data from frame-based input. It uses an advanced pixel model that mimics the dynamic properties of DVS sensors, including noise, polarity events, latency, and contrast sensitivity. Moreover, it incorporates Super-SloMo interpolation, which increases the temporal resolution by generating intermediate frames—crucial for detecting fast movements like falls. This capability allows researchers to work with virtual DVS sequences that maintain the temporal and structural richness required by spiking architectures detection.
The specific methodology can be broken into four interconnected stages, (1) Design of the hybrid neural architecture, combining spatial processing through CNNs and temporal spike processing via SNNs, (2) Creation of a custom dataset (NFDD) using RGB video transformed into neuromorphic event data through v2e, (3) Data augmentation techniques to increase dataset size and variety without additional recordings, and (4) Training and validation of the model using surrogate gradient methods adapted for spiking neurons.

\section{Implementation}
This section details the methodological process followed for the design, training, and evaluation of hybrid models combining Convolutional Neural Networks (CNNs) with Spiking Neural Networks (SNNs). All results obtained from the different architectures presented below will be discussed in the chapter Results.

\subsection{Data Collection and Event Simulation using v2e Dataset}
\begin{figure}[t]
\centering
\includegraphics[width=\linewidth]{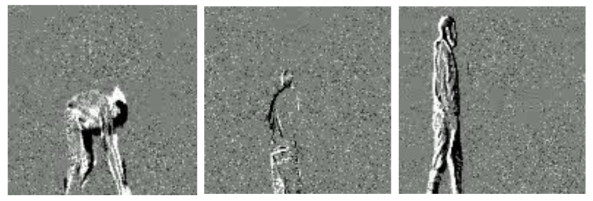}
\caption{Falling, sitting and walking classes transformed into events.}
\label{fig:4}
\end{figure}

The starting point was the capture of video data using a conventional mobile phone camera, configured to record at a resolution of 1280×720 pixels. To ensure the quality and focus of the data, the camera was kept fixed on a tripod, avoiding any scene motion that could introduce unwanted background activity into the dataset. Videos were recorded in controlled indoor environments under various lighting conditions to test the model's robustness. Each video lasted five seconds, and three activity classes were recorded: Walking, Sitting, and Falling (Fig.\ref{fig:4}).

Once collected, the raw videos were processed using v2e to simulate the output of an event-based camera. As mentioned, V2e converts traditional frame-based video into asynchronous event streams by analyzing pixel-wise changes in luminance and generating events based on thresholds of contrast change. It replicates the behavior of real DVS sensors, including non-idealities such as sensor noise and temporal jitter. This makes the simulated data nearly indistinguishable from real sensor outputs.
An optional but highly impactful feature used during this step was the Super-SloMo integration. By interpolating intermediate frames from the original video, the v2e tool effectively increased the temporal resolution of the event streams. This was especially beneficial for capturing short, high-speed motions—such as the moment a fall begins—allowing the model to detect patterns that might otherwise be lost in conventional video processing.

\subsection{Dataset Construction and Augmentation}
The resulting dataset, named the Neuromorphic Falling Detection Dataset (NFDD), was constructed to serve as a representative and well-balanced collection of neuromorphic data. Initially, 100 samples per class were recorded (300 total). To enhance dataset diversity and model generalization, a multi-step data augmentation pipeline was applied: Horizontal flipping doubled the dataset to 600 samples and Zoom-in operations and minor rotations further expanded the dataset to 1,200 samples total (400 per class).
 Each sample was resized from the original 1280×720 resolution to a standard 128×128 pixel format, optimizing memory usage and processing speed while retaining key spatial information. This made the dataset manageable for training on hybrid networks while preserving the fidelity of the events.
Compared to the DVSGesture dataset, which was used as an initial benchmark, NFDD provides fewer samples per class but with more relevant, domain-specific activities. Additionally, the event data produced through v2e allows for better temporal modeling, which is particularly important for safety-critical actions like falls.

\subsection{Hybrid CNN + SNN Network Architecture}
The architecture used for classification builds upon the model previously tested on DVSGesture (Fig.\ref{fig:1}), but it includes several refinements. 
In this original implementation, the CNN performs downsampling and convolution operations to reduce the spatial dimensions from 128×128 to 32×32 pixels, as mentioned, enabling efficient processing while retaining discriminative features. The resulting features are passed into the SNN layers, where temporal dynamics are modeled using leaky integrate-and-fire (LIF) neurons. A rate coding scheme was applied, translating event densities over small time windows into spike representations.
Training was carried out using surrogate gradient descent, an adaptation that allows for the backpropagation of gradients through non-differentiable spiking functions. The Adam optimizer was employed with a batch size of 32 and a learning rate tuned empirically.

\section{Results and Analysis}
\begin{figure}[t]
\centering
\includegraphics[width=\linewidth]{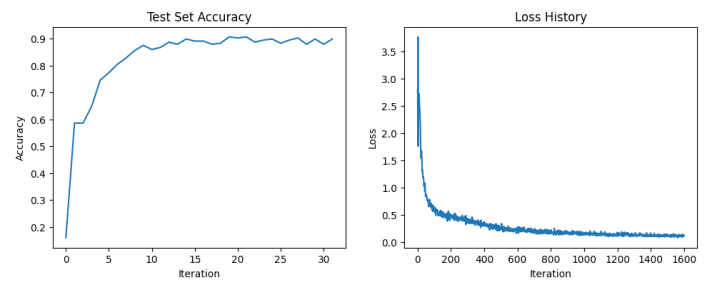}
\caption{Training accuracy and loss history of DVS128 after 100 epochs.}
\label{fig:5}
\end{figure}

\begin{figure}[t]
\centering
\includegraphics[width=\linewidth]{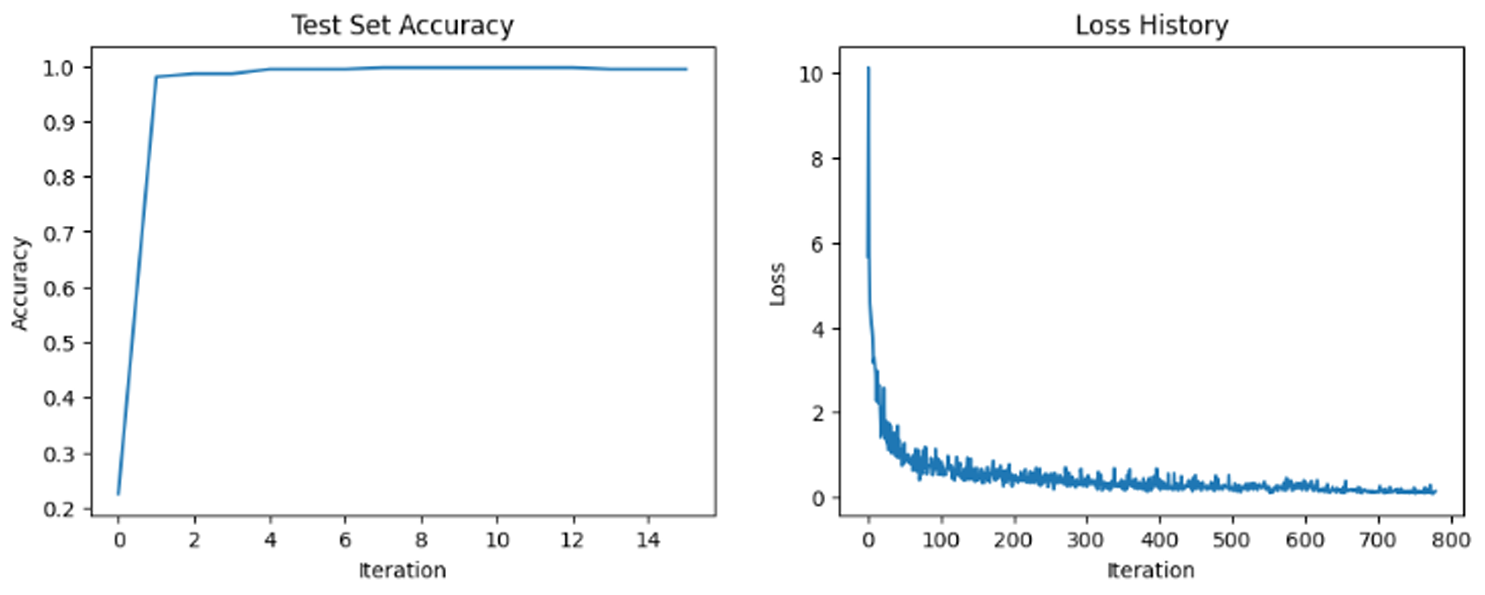}
\caption{Training and validation of the Neural Network with NFDD dataset.}
\label{fig:6}
\end{figure}

Training the hybrid CNN-SNN model on the NFDD dataset produced good results. After applying data augmentation techniques and configuring the neural network architecture without downsampling the spatial resolution (preserving 128×128 pixels), the model achieved a final classification accuracy of 99.72\%.
The significantly high accuracy highlights several key aspects. The volume and diversity of the NFDD dataset, expanded through augmentation (flipping, zooming, and rotation), allowed the model to generalize better without overfitting. Capturing videos under controlled static conditions minimized irrelevant background noise, allowing the network to focus exclusively on the relevant action dynamics (walking, sitting, and falling). Maintaining the original input resolution enabled the network to extract richer spatial features across the event data streams.
Fig.\ref{fig:5} and Fig.\ref{fig:6} show the training and validation accuracy curves over the 100 epochs, indicating stable convergence and minimal variance between training and validation, confirming the robustness of the model.
Moreover, by simulating event data from standard RGB videos using the v2e library, this study demonstrated that effective neuromorphic datasets can be generated without requiring specialized dynamic vision sensors, significantly reducing hardware dependency and cost.

\section{Discussion}
This paper presented a framework for training fall detection models using synthetic event-based data. To achieve this, a hybrid CNN-SNN architecture was proposed and evaluated using both a public DVS Gesture dataset and a custom fall detection dataset called NFDD. The use of the v2e simulator libraries proved to be an effective and low-cost method for generating training data from standard smartphone cameras. The proposed model achieved 91.7\% accuracy on the DVS Gesture dataset and 99.7\% accuracy on the NFDD dataset. Future work includes extending the classification of fall dynamics based on biomechanical models, developing a fully SNN-based implementation of the architecture, and deploying the models on low-power hardware, including neuromorphic platforms. 

\section*{Acknowledgment}
Acknowledgment to the Chilean National Research and Development Agency (ANID) for support through the Fondecyt Iniciación Project No. 11251536.

\bibliographystyle{IEEEtran}
\bibliography{references}

@article{gallego2020event,
  author={Gallego, Guillermo and Delbruck, Tobi and Orchard, Garrick and Bartolozzi, Chiara and Taba, Bahadir and Censi, Andrea and Leutenegger, Stefan and Davison, Andrew J. and Conradt, Jörg and Daniilidis, Kostas and Scaramuzza, Davide},
  title={Event-based Vision: A Survey},
  journal={IEEE Transactions on Pattern Analysis and Machine Intelligence},
  volume={44},
  number={1},
  pages={154--180},
  year={2020}
}

@article{cicirelli2021ambient,
  author={Cicirelli, Grazia and Marani, Roberto and Petitti, Antonio and Milella, Antonio and D'Orazio, Tommaso},
  title={Ambient Assisted Living: A Review of Technologies, Methodologies and Future Perspectives for Healthy Aging of Population},
  journal={Sensors},
  volume={21},
  number={10},
  pages={3549},
  year={2021}
}

@article{wang2020elderly,
  author={Wang, Xueyi and Ellul, Joshua and Azzopardi, George},
  title={Elderly Fall Detection Systems: A Literature Survey},
  journal={Frontiers in Robotics and AI},
  volume={7},
  pages={71},
  year={2020}
}

@article{singh2020sensor,
  author={Singh, Anuradha and others},
  title={Sensor Technologies for Fall Detection Systems: A Review},
  journal={IEEE Sensors Journal},
  volume={20},
  number={13},
  pages={6889--6919},
  year={2020}
}

@inproceedings{wang2023fall,
  author={Wang, Xueyi and others},
  title={Fall Detection with Event-Based Data: A Case Study},
  booktitle={International Conference on Computer Analysis of Images and Patterns},
  publisher={Springer Nature Switzerland},
  year={2023}
}

@inproceedings{hu2021v2e,
  author={Hu, Yuhuang and Liu, Shih-Chii and Delbruck, Tobi},
  title={v2e: From Video Frames to Realistic DVS Events},
  booktitle={Proceedings of the IEEE/CVF Conference on Computer Vision and Pattern Recognition},
  year={2021}
}

@inproceedings{prasad2023hybrid,
  author={Prasad, Shyam Sunder and others},
  title={Hybrid SNN-Based Privacy-Preserving Fall Detection Using Neuromorphic Sensors},
  booktitle={Proceedings of the Fourteenth Indian Conference on Computer Vision, Graphics and Image Processing},
  year={2023}
}

@article{lee2017embedded,
  author={Lee, Hyunwoo and others},
  title={Embedded Real-Time Fall Detection Using Deep Learning for Elderly Care},
  journal={arXiv preprint arXiv:1711.11200},
  year={2017}
}

@inproceedings{krishnan2022benchmarking,
  author={Krishnan, Karthik Sivarama and Krishnan, Koushik Sivarama},
  title={Benchmarking Conventional Vision Models on Neuromorphic Fall Detection and Action Recognition Dataset},
  booktitle={2022 IEEE 12th Annual Computing and Communication Workshop and Conference (CCWC)},
  publisher={IEEE},
  year={2022}
}

@inproceedings{belbachir2012care,
  author={Belbachir, Ahmed Nabil and others},
  title={CARE: A Dynamic Stereo Vision Sensor System for Fall Detection},
  booktitle={2012 IEEE International Symposium on Circuits and Systems (ISCAS)},
  publisher={IEEE},
  year={2012}
}

@misc{nfdd2024,
  author={Rojas, Guillermo},
  title={NFDD\_SNN Kaggle Notebook},
  year={2024},
  note={[Online]. Available: https://www.kaggle.com/code/apemangr/nfdd-snn}
}

@inproceedings{amir2017low,
  author={Amir, Arnon and others},
  title={A Low Power, Fully Event-Based Gesture Recognition System},
  booktitle={Proceedings of the IEEE Conference on Computer Vision and Pattern Recognition},
  year={2017}
}

\end{document}